%% file: egpaper_for_review.tex
\documentclass[10pt,twocolumn,letterpaper]{article}

\usepackage{iccv}
\usepackage{times}
\usepackage{epsfig}
\usepackage{graphicx}
\usepackage{amsmath}
\usepackage{amssymb}

\usepackage{float}
\usepackage[utf8]{inputenc}
\usepackage{makeidx}
\usepackage{url}
\usepackage{epstopdf}
\usepackage{doc}
\usepackage{enumerate}
\usepackage{bbm}
\usepackage{multicol, blindtext}
\usepackage{ctable}
\usepackage{textcomp}			
\usepackage[singlelinecheck=false]{caption}
\usepackage[]{footmisc}
\usepackage{multirow}

\usepackage[font={small}]{caption}
\iccvfinalcopy 

\def\iccvPaperID{11} 

\begin{document}

\title{Resource Efficient 3D Convolutional Neural Networks}

\author{\parbox{16cm}{\centering
    {\large Okan K\"op\"ukl\"u$^1$, Neslihan Kose$^2$, Ahmet Gunduz$^1$, Gerhard Rigoll$^1$}\\
    {\normalsize
    \vspace{0.2cm}
    $^1$ Institute for Human-Machine Communication, TU Munich, Germany\\
    $^2$ Dependability Research Lab, Intel Labs Europe, Intel Deutschland GmbH, Germany}}
}

\maketitle

\input{Abstract.tex}
\input{Introduction.tex}

\input{Related_Work.tex}
\input{Method.tex}

\input{Experiments.tex}

\input{Conclusion.tex}
\input{Acknowledgements.tex}

{\small
\bibliographystyle{ieee}
\bibliography{egbib}
}

\end{document}

%% file: Abstract.tex
\begin{abstract}

Recently, convolutional neural networks with 3D kernels (3D CNNs) have been very popular in computer vision community as a result of their superior ability of extracting spatio-temporal features within video frames compared to 2D CNNs. Although there has been great advances recently to build resource efficient 2D CNN architectures considering memory and power budget, there is hardly any similar resource efficient architectures for 3D CNNs. In this paper, we have converted various well-known resource efficient 2D CNNs to 3D CNNs and evaluated their performance on three major benchmarks in terms of classification accuracy for different complexity levels. We have experimented on (1) Kinetics-600 dataset to inspect their capacity to learn, (2) Jester dataset to inspect their ability to capture motion patterns, and (3) UCF-101 to inspect the applicability of transfer learning. We have evaluated the run-time performance of each model on a single Titan XP GPU and a Jetson TX2 embedded system. The results of this study show that these models can be utilized for different types of real-world applications since they provide real-time performance with considerable accuracies and memory usage. Our analysis on different complexity levels shows that the resource efficient 3D CNNs should not be designed too shallow or narrow in order to save complexity. The codes and pretrained models used in this work are publicly available \footnote{https://github.com/okankop/Efficient-3DCNNs}.

\end{abstract}

%% file: Introduction.tex
\section{Introduction}

Ever since AlexNet \cite{krizhevsky2012imagenet} won the ImageNet Challenge (ILSVRC 2012 \cite{russakovsky2015imagenet}), convolutional neural networks (CNNs) have dominated the majority of the computer vision tasks. Then the primary trend has been more on creating deeper and wider CNN architectures to achieve higher accuracies \cite{he2016deep, simonyan2014very, szegedy2015going}. However, in real world computer vision applications such as face recognition, robot navigation and  augmented reality, the tasks need to be carried out under runtime constraints on a computationally limited platform. Only recently, there has been a rising interest in building resource efficient convolutional neural networks but it is limited with 2-dimensional kernels (2D) \cite{iandola2016squeezenet, howard2017mobilenets, Zhang2018ShuffleNetAE, ma2018shufflenet, sandler2018mobilenetv2}.

The same history is repeating for CNNs with 3-dimensional (3D) kernels \cite{hara2018can}. Since the large video datasets became available, the primary trend for video recognition tasks is again to achieve higher accuracies by building deeper and wider architectures \cite{tran2017convnet, qiu2017learning, tran2018closer, hara2018can, feichtenhofer2018slowfast}. Considering the fact that 3D CNNs achieve better performance for video recognition tasks compared to 2D CNNs \cite{carreira2017quo}, it is very likely that this 3D CNN architecture search will continue until the achieved accuracies saturate. However, real-world applications still require resource efficient 3D CNN architectures taking runtime, memory and power budget into account. This work aims to fill this research gap.

In this paper, we first have created the 3D versions of the well-known 2D resource efficient architectures: SqueezeNet, MobileNet, ShuffleNet, MobileNetV2 and ShuffleNetV2. We have evaluated t-he performance of these architectures on three publicly available benchmarks:

\begin{enumerate}[(1)]
\item Kinetics-600 dataset\cite{carreira2017quo} to learn models' capacities.
\item Jester dataset \cite{jester} to learn how well the models capture the motion. 
\item UCF-101 dataset \cite{soomro2012ucf101} to evaluate the applicability of transfer learning for each model.
\end{enumerate}

The computational complexity of the implemented architectures are measured in terms of floating point operations (FLOPs), which is widely used metric among resource efficient architectures. In this paper, the number of FLOPs refers to the number of multiply-adds. However, as highlighted by \cite{ma2018shufflenet}, the number of FLOPs is an indirect metric which does not give an actual performance indication like speed or latency. Therefore, for all the implemented architectures we have also evaluated their run-time performance on two different platforms, which are Nvidia Titan XP GPU and Jetson TX2 embedded system-on-module (SoM) with integrated 256-core Pascal GPU.



%% file: Related_Work.tex
\section{Related Work}

Lately, there is a rising interest in building small and efficient neural networks \cite{iandola2016squeezenet, howard2017mobilenets, ma2018shufflenet, rastegari2016xnor, wu2016quantized, han2015deep}. The common approaches used for this objective can be categorized under two categories: (i) Accelerating the pretrained networks, or (ii) directly constructing small networks by manipulating kernels. For the first one, \cite{han2015deep, han2015learning, wen2016learning, molchanov2016pruning} proposes to prune either network connections or channels without reducing the performance of pretrained models. Additionally, many other methods apply quantization \cite{rastegari2016xnor, soudry2014expectation, wu2016quantized} or factorization \cite{lebedev2014speeding, jaderberg2014speeding, jin2014flattened} for the same objective. However, our focus is on the second one for directly designing small and resource efficient 3D CNN architectures.

Current well-known resource efficient CNN architectures are all constructed with 2D convolutional kernels and benchmarked at ImageNet. SqueezeNet \cite{iandola2016squeezenet} reduced the number of parameters and computation while maintaining the classification performance. MobileNet \cite{howard2017mobilenets} makes use of depthwise separable convolutions to construct light-weight deep neural networks. The depthwise separable convolutions factorize the standard convolutions into a depthwise convolution followed by a 1x1 pointwise convolution. Compared to standard convolutions, depthwise separable convolutions use between 8 to 9 times less parameters and computations. ShuffleNet \cite{Zhang2018ShuffleNetAE} proposes to use pointwise group convolutions and channel shuffle in order to reduce computational cost. MobileNetv2 \cite{sandler2018mobilenetv2} makes use of the inverted residual structure where the intermediate expansion layer uses depthwise convolutions. ShuffleNetV2 \cite{ma2018shufflenet} builds on top of ShuffleNet \cite{Zhang2018ShuffleNetAE} using channel split together with channel shuffle which realizes a feature reuse pattern.    

These architectures intensively make use of group convolutions and depthwise separable convolutions. Group convolutions are first introduced in AlexNet \cite{krizhevsky2012imagenet} and efficiently utilized in ResNeXt \cite{xie2017aggregated}. Depthwise separable convolutions are introduced in Xception \cite{chollet2017xception} and they are the main building blocks for majority of lightweight architectures. 

All of the above-mentioned resource efficient architectures are 2D CNNs. They are designed to operate on static images and evaluated on a very large benchmark (i.e., ImageNet). To the best of our knowledge, this is the first work that proposes and evaluates resource efficient 3D CNNs on large scale video benchmarks. 

3D CNNs such as well-known C3D \cite{tran2015learning} require significantly more parameters and computations compared to their 2D counterparts which make them harder to train and prone to overfitting. With the availability of large scale video datasets such as Sports-1M \cite{karpathy2014large}, Kinetics-400 \cite{carreira2017quo}, this problem is solved. Moreover, \cite{carreira2017quo} proved that 3D CNNs achieve better accuracies compared to 2D CNNs for video classification task. Consequently, 3D CNN architecture search is an active area in research community to achieve higher accuracies. 

Several 3D CNN architectures have been proposed recently. Carreira et al. propose Inflated 3D CNN (I3D) \cite{carreira2017quo}, where the filters and pooling kernels of a deep CNN are expanded to 3D, making it possible to leverage successful ImageNet architecture designs and their pretrained models. P3D \cite{qiu2017learning} and (2+1)D \cite{tran2018closer} propose to decompose 3D convolutions into  2D and 1D convolutions operating on spatial and depth dimensions, respectively. In \cite{hara2018can}, 3D versions of famous ImageNet architectures such as ResNet \cite{he2016deep}, Wide ResNet \cite{zagoruyko2016wide}, ResNeXt \cite{xie2017aggregated} and DenseNet \cite{huang2017densely} are evaluated and it has been shown that ResNeXt achieves better results compared to others. Recently, Feichtenhofer et al. propose a novel architecture named SlowFast \cite{feichtenhofer2018slowfast}, which uses a Slow pathway, operating at low frame rate, to capture static content of a video, and a Fast pathway, operating at high frame rate, to capture the dynamic content of a video.

Up to now, nearly all the 3D CNN architectures in the literature are heavyweight, requiring 10s and even 100s billions of floating point operations (FLOPs). Moreover, majority of these architectures also use optical flow modality, which increases the complexity even further. Our focus in this work is to evaluate 3D CNNs having less than 1 GFLOPs. Consequently, we have implemented the 3D version of SqueezeNet \cite{iandola2016squeezenet}, MobileNet \cite{howard2017mobilenets}, MobileNetV2 \cite{sandler2018mobilenetv2}, ShuffleNet \cite{Zhang2018ShuffleNetAE} and ShuffleNetV2 \cite{ma2018shufflenet} for 4 different complexity levels and then evaluated them on 3 different video benchmarks.  We have evaluated our architectures only using RGB modality without computing costly optical flow modality. 



%% file: Method.tex
\section{Resource Efficient 3D CNN Architectures}

\begin{figure*}[t!]
	\centering
	\includegraphics[width=1.0\textwidth]{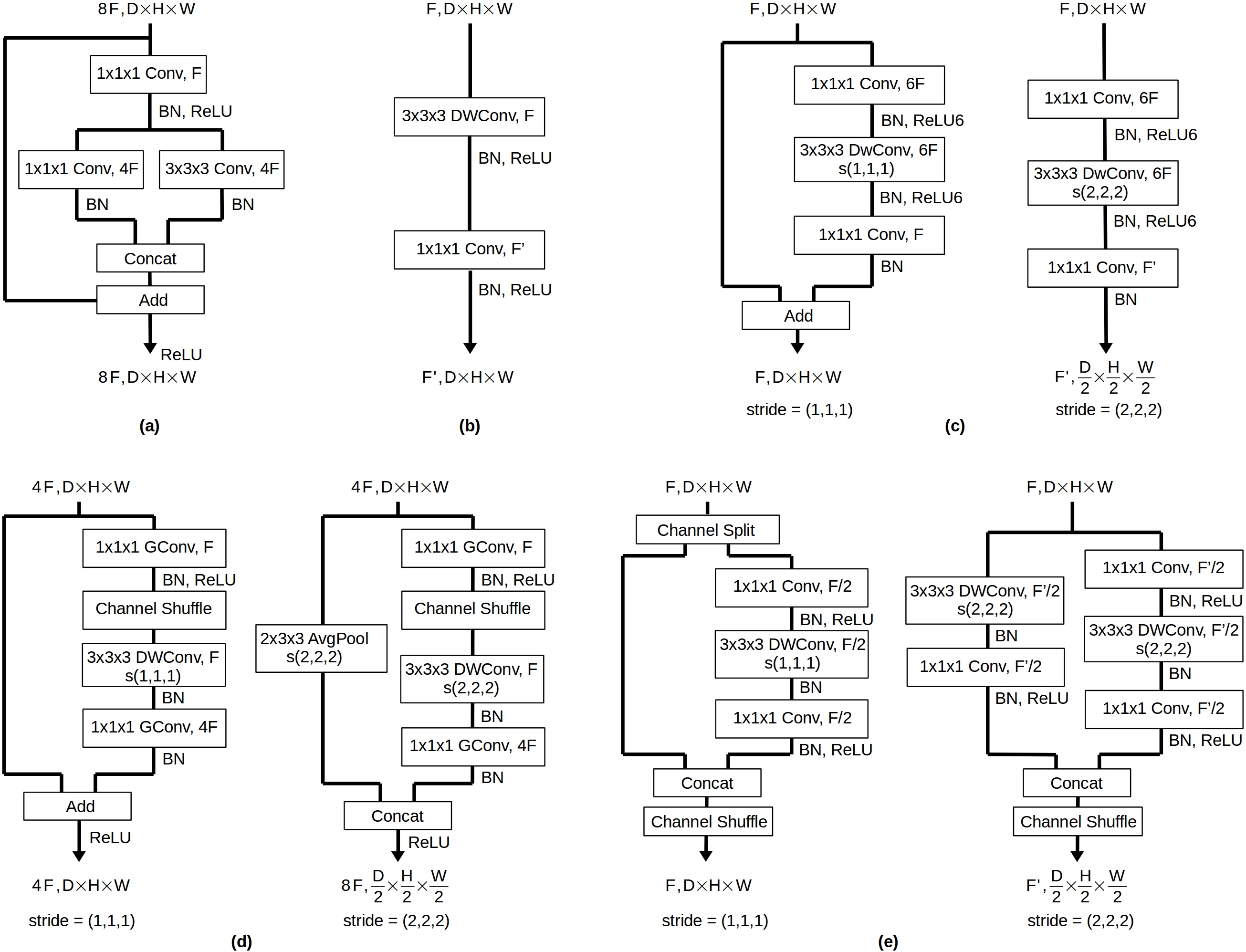}
	\caption{Main building block for each resource efficient 3D CNN architecture. \textit{F} is the number of feature maps and D $\times$ H $\times$ W stands for Depth $\times$ Height $\times$ Width for the input and output volumes.  \textit{DWConv} and \textit{GConv} stand for depthwise and group convolution, respectively. \textit{BN} and \textit{ReLU(6)} stand for Batch Normalization and Rectified Linear Unit (capped at 6), respectively. \textbf{(a)} SqueezeNet's Fire block; \textbf{(b)} MobileNet block; \textbf{(c)} left: MobileNetv2 block, right: MobileNetv2 block with spatiotemporal downsampling (2x); \textbf{(d)} left: ShuffleNet block, right: ShuffleNet block with spatiotemporal downsampling (2x); \textbf{(e)} left: ShuffleNetv2 block, right: ShuffleNetv2 block with spatiotemporal downsampling (2x).}
	\label{fig:blocks}
\end{figure*}

In this section, we explain the details of the resource efficient 3D CNN architectures that have been proposed and evaluated within the scope of this work. We initially introduce the 3D versions of the well-know resource efficient 2D CNN architectures by explaining their building blocks and networks structures. Then we compare these models in terms of number of layers, nonlinearities, and skip connections. We conclude with training details of the models. 


\subsection{3D Versions of Well-known Architectures}

In this section, we give the implementation details of our resource efficient architectures with 3-dimensional kernels, which are converted from well-know resource efficient 2D CNN architectures. 
Main building blocks of each architecture are depicted in Fig. \ref{fig:blocks}. The input is always considered as a clip of 16 frames with spatial resolution of 112 pixels. For all of the 3D CNN architectures, first convolutions always apply stride of (1,2,2). For the rest of the architectures, depth dimension is reduced together with spatial dimensions. 

\subsubsection{3D-SqueezeNet}

\begin{table}[t!]
	\centering
	\begin{tabular}{lll}
		\specialrule{.15em}{.0em}{.2em}
		\textbf{Layer / Stride}    & \textbf{Filter size}  & \textbf{Output size}\\ 
		\specialrule{.15em}{.2em}{.2em}
		Input clip            &         & 3x16x112x112  \\
		Conv1/s(1,2,2)        & 3x3x3   & 64x16x56x56   \\
		MaxPool/s(2,2,2)      & 3x3x3   & 64x8x28x28    \\
		\specialrule{.1em}{.1em}{.1em}
		Fire2                 &         & 128x8x28x28   \\
		Fire3                 &         & 128x8x28x28   \\
		MaxPool/s(2,2,2)      & 3x3x3   & 128x4x14x14   \\
		\specialrule{.1em}{.1em}{.1em}
		Fire4                 &         & 256x4x14x14   \\
		Fire5                 &         & 256x4x14x14   \\
		MaxPool/s(2,2,2)      & 3x3x3   & 256x2x7x7     \\
		\specialrule{.1em}{.1em}{.1em}
		Fire6                 &         & 384x2x7x7     \\
		Fire7                 &         & 384x2x7x7     \\
		MaxPool/s(2,2,2)      & 3x3x3   & 384x1x4x4     \\
		\specialrule{.1em}{.1em}{.1em}
		Fire8                 &         & 512x1x4x4     \\
		Fire9                 &         & 512x1x4x4     \\
		\specialrule{.1em}{.1em}{.1em}
		Conv10/s(1,1,1)       & 1x1x1   & \textit{NumCls}x1x4x4   \\
		AvgPool/s(1,1,1)      & 1x4x4   & \textit{NumCls} \\
		\specialrule{.15em}{.2em}{.0em}
	\end{tabular}
	\caption{3D-SqueezeNet architecture. Details of Fire block is given in Fig. \ref{fig:blocks} (a).}
	\label{tab:squeezenet_arch}
\end{table}

SqueezeNet \cite{iandola2016squeezenet} is considered as one of very first resource efficient CNN architectures with notable accuracy performance. It achieves the AlexNet \cite{krizhevsky2012imagenet}-level accuracy with 50 times fewer parameters and less than 0.5 MB model size. 

The main building block of SqueezeNet is Fire block whose 3D version is depicted in Fig. \ref{fig:blocks} (a). As illustrated in Table \ref{tab:squeezenet_arch}, 3D-SqueezeNet begins with a convolution layer (Conv1), followed by 8 Fire blocks (Fire-2-9), ending with a final convolutional layer (Conv10). 

In our experiments, we use SqueezeNet with simple bypass since it achieves the best result in its 2D version for ImageNet. SqueezeNet does not apply depthwise convolutions which is the main building block for majority of resource efficient architectures. Instead, it uses three strategies to reduce the number of parameters while maintaining accuracy: (i) Replacing 3x3 filters with 1x1 filters, (ii) decreasing the number of input channels to 3x3 filters, and (iii) downsampling late in the network so that convolution layers have large activation maps. Moreover, compared to other resource efficient architectures, SqueezeNet cannot be modified with $width\_multiplier$ parameter resulting in different complexities. Therefore, it is only experimented with its default configuration as shown in Table \ref{tab:comparison}.

\subsubsection{3D-MobileNetV1}
MobileNets \cite{howard2017mobilenets} apply depthwise separable convolutions which have a form that factorize a standard convolution into a depthwise convolution and $1\times1$ convolution, which is called as pointwise convolution. In MobileNet architectures, the depthwise convolution applies a single filter to each input channel and then the pointwise convolution applies a $1\times1$ convolution to combine the outputs of the depthwise convolution. Different from the standard convolution, the depthwise separable convolution involves two layers which separates filtering and combining operations as illustrated in Fig. \ref{fig:blocks} (b). This process helps to decrease computation time and model size significantly. Unlike all recent popular CNN architectures, MobileNet does not contain skip connections. Therefore, depth of the network cannot be increased too much which hinders gradient flow.

Table \ref{tab:mobilenet_arch} shows the details of the 3D-MobileNet architecture. 3D-MobileNet begins with a convolutional layer, followed by 13 MobileNet blocks, ending with a linear layer. MobileNet has 28 layers in case the depthwise and pointwise convolutions in each MobileNet block are counted as separate layers. 

\begin{table}[t!]
	\centering
	\begin{tabular}{p{3cm}cl}
		\specialrule{.15em}{.0em}{.2em}
		\textbf{Layer / Stride}    & \textbf{Repeat}  & \textbf{Output size}\\ 
		\specialrule{.15em}{.2em}{.1em}
		Input clip              &          & 3x16x112x112  \\
		Conv(3x3x3)/s(1,2,2)    & 1        & 32x16x56x56   \\
		\specialrule{.1em}{.1em}{.1em}
		Block/s(2x2x2)          & 1        & 64x8x28x28    \\
		Block/s(2x2x2)          & 1        & 128x4x14x14   \\
		Block/s(1x1x1)          & 1        & 128x4x14x14   \\
		Block/s(2x2x2)          & 1        & 256x2x7x7     \\
		Block/s(1x1x1)          & 1        & 256x2x7x7     \\
		Block/s(2x2x2)          & 1        & 512x1x4x4     \\
		Block/s(1x1x1)          & 5        & 512x1x4x4     \\
		Block/s(1x1x1)          & 1        & 1024x1x4x4    \\
		Block/s(1x1x1)          & 1        & 1024x1x4x4    \\
		\specialrule{.1em}{.1em}{.1em}
		AvgPool(1x4x4)/s(1,1,1)     & 1   & 1024x1x1x1     \\
		Linear(1024x\textit{NumCls})  & 1 & \textit{NumCls} \\
		\specialrule{.15em}{.2em}{.1em}
	\end{tabular}
	\caption{3D-MobileNet architecture. Details of Block is given in Fig. \ref{fig:blocks} (b).}
	\label{tab:mobilenet_arch}
\end{table}

\subsubsection{3D-MobileNetV2}

\begin{table}[b!]
	\centering
	\begin{tabular}{p{2.6cm}cl}
		\specialrule{.15em}{.0em}{.2em}
		\textbf{Layer / Stride}    & \textbf{Repeat}  & \textbf{Output size}\\ 
		\specialrule{.15em}{.2em}{.1em}
		Input clip              &      & 3x16x112x112    \\
		Conv(3x3x3)/s(1,2,2)    & 1    & 32x16x56x56     \\
		\specialrule{.1em}{.1em}{.1em}
		Block/s(1x1x1)          & 1    & 16x16x56x56     \\
		Block/s(2x2x2)          & 2    & 24x8x28x28      \\
		Block/s(2x2x2)          & 3    & 32x4x14x14      \\
		Block/s(2x2x2)          & 4    & 64x2x7x7        \\
		Block/s(1x1x1)          & 3    & 96x2x7x7        \\
		Block/s(2x2x2)          & 3    & 160x1x4x4       \\
		Block/s(1x1x1)          & 1    & 320x1x4x4       \\
		\specialrule{.1em}{.1em}{.1em}
		Conv(1x1x1)/s(1,1,1)    & 1    & 1280x1x4x4      \\
		AvgPool/s(1,1,1)        & 1    & 1024x1x1x1      \\
		Linear                  & 1    & \textit{NumCls} \\
		\specialrule{.15em}{.2em}{.1em}
	\end{tabular}
	\caption{3D-MobileNetV2 architecture. Block is inverted residual block whose details are given in Fig. \ref{fig:blocks} (c) with stride 1 (left) and spatio temporal 2x downsampling (right).}
	\label{tab:mobilenetv2_arch}
\end{table}

MobileNetV2 \cite{sandler2018mobilenetv2} is another 2D resource efficient architecture. It builds upon the main idea of MobileNetV1 by using depthwise separable convolutions; however, it introduces two new components: 1) linear bottlenecks between the layers, and 2) shortcut connections between the bottlenecks. The idea behind 1) is both keeping the size of model low by decreasing number of channels and extracting as much as information by applying depthwise convolution after decompressing the data. This convolutional module allows to reduce memory usage during inference. On the other hand, 2) allows training faster and construct deeper models like ResNet architectures \cite{he2016deep}. 

Fig. \ref{fig:blocks} (c) shows the MobileNetV2 block. Table \ref{tab:mobilenetv2_arch} shows the layers of 3D-MobileNetV2 architecture. 3D-MobileNetV2 begins with a convolutional layer, followed by 17 MobileNetV2 blocks, and then a convolutional layer and finally ending with a linear layer.

\subsubsection{3D-ShuffleNetV1}
According to \cite{Zhang2018ShuffleNetAE}, ShuffleNet provides superior performance compared to MobileNet \cite{howard2017mobilenets} by a significant margin, which is reported as absolute 7.8\% lower ImageNet top-1 error at level of 40 MFLOPs. The model is also reported to achieve ~$13 \times$ actual speedup over AlexNet while maintaining comparable accuracy.

The architecture uses two new operations, which are pointwise group convolution and channel shuffle which is depicted in Fig. \ref{fig:blocks} (d). 

As illustrated in Table \ref{tab:shufflenet_arch}, 3D-ShuffleNet begins with a convolutional layer followed by 16 ShuffleNet blocks, which are grouped into three stages. In each stage, the number of output channels are kept same with the applied ShuffleNet blocks. For the next stage, the output channels are doubled and the spatial and depth dimensions are reduced to half. ShuffleNet architecture ends with a final linear layer. In ShuffleNet units, group number $g$ controls the connection sparsity of pointwise convolutions. In this study, the group number is selected as 3.

\begin{table}[t!]
	\centering
	\begin{tabular}{p{3cm}cl}
		\specialrule{.15em}{.0em}{.2em}
		\textbf{Layer / Stride}    & \textbf{Repeat}  & \begin{tabular}[c]{@{}c@{}}\textbf{Output size}\\\textbf{(groups=3)}\end{tabular}\\ 
		\specialrule{.15em}{.2em}{.1em}
		Input clip              &      & 3x16x112x112       \\
		Conv(3x3x3)/s(1,2,2)    & 1    & 24x16x56x56     \\
		MaxPool(3x3x3)/s(2,2,2) & 1    & 24x8x28x28    \\
		\specialrule{.1em}{.1em}{.1em}
		Block/s(2x2x2)          & 1    & 240x4x14x14     \\
		Block/s(1x1x1)          & 3    & 240x4x14x14      \\
		\specialrule{.1em}{.1em}{.1em}
		Block/s(2x2x2)          & 1    & 480x2x7x7      \\
		Block/s(1x1x1)          & 7    & 480x2x7x7        \\
		\specialrule{.1em}{.1em}{.1em}
		Block/s(2x2x2)          & 1    & 960x1x4x4        \\
		Block/s(1x1x1)          & 3    & 960x1x4x4       \\
		\specialrule{.1em}{.1em}{.1em}
		AvgPool(1x4x4)/s(1,1,1) & 1    & 960x1x1x1      \\
		Linear                  & 1    & \textit{NumCls} \\
		\specialrule{.15em}{.2em}{.1em}
	\end{tabular}
	\caption{3D-ShuffleNet architecture.  Its' main building block is given in Fig. \ref{fig:blocks} (d) with stride 1 (left) and spatio temporal 2x downsampling (right).}
	\label{tab:shufflenet_arch}
\end{table}

\subsubsection{3D-ShuffleNetV2}

In ShuffleNetV2 \cite{ma2018shufflenet} architecture, channel split operator is introduced different from V1. As illustrated in Fig. \ref{fig:blocks} (e), at the beginning of each block, the input of \textit{c} feature channels are split into two branches with c-c$^{'}$ and $c^{'}$ channels, respectively. One branch remains as identity, and the other branch includes three convolutions with the same input and output channels. Different from ShuffleNet, the two 1$\times$1 convolutions are not groupwise. After the convolutions, the two branches are concatenated and the number of channels keeps the same. At the end of the block, channel shuffle operation is applied to enable information communication between the two branches.

Table \ref{tab:shufflenetv2_arch} shows the layers of 3D-ShuffleNetV2 architecture. 3D-ShuffleNetV2 architecture begins with a convolutional layer, followed by 16 ShuffleNetV2 blocks, and then a convolutional layer and finally ending with a linear layer. Similar to 3D-ShuffleNet, the stack of blocks are grouped into three stages, and at each stage the number of output channels are kept same while with the next stage, they are doubled. Different from the 3D-ShuffleNet, the number of channels in each stage are not fixed. Table \ref{tab:shufflenetv2_ch} shows the number of channels ($c_1$, $c_2$, $c_3$, $c_4$) for different levels of complexities. Also, in 3D-ShuffleNet, the number of output channels in the final layer ($c_4$) is same after the third stage, whereas in 3D-ShuffleNetV2, different number of output channels are selected for different levels of complexities (Table \ref{tab:shufflenetv2_ch}).

\begin{table}[t!]
	\centering
	\begin{tabular}{p{3cm}clc}
		\specialrule{.15em}{.0em}{.2em}
		\textbf{Layer / Stride}    & \textbf{Repeat}  & \textbf{Output size}  \\ 
		\specialrule{.15em}{.2em}{.1em}
		Input clip              &      & 3x16x112x112    \\
		Conv(3x3x3)/s(1,2,2)    & 1    & 24x16x56x56     \\
		MaxPool(3x3x3)/s(2,2,2) & 1    & 24x8x28x28      \\
		\specialrule{.1em}{.1em}{.1em}
		Block/s(2x2x2)          & 1    & $c_1$x4x14x14  \\
		Block/s(1x1x1)          & 3    & $c_1$x4x14x14   \\
		\specialrule{.1em}{.1em}{.1em}
		Block/s(2x2x2)          & 1    & $c_2$x2x7x7   \\
		Block/s(1x1x1)          & 7    & $c_2$x2x7x7     \\
		\specialrule{.1em}{.1em}{.1em}
		Block/s(2x2x2)          & 1    & $c_3$x1x4x4     \\
		Block/s(1x1x1)          & 3    & $c_3$x1x4x4     \\
		\specialrule{.1em}{.1em}{.1em}
		Conv(1x1x1)/s(1,1,1)    & 1    & $c_4$x1x4x4     \\
		AvgPool(1x4x4)/s(1,1,1) & 1    & $c_4$x1x1x1     \\
		Linear                  & 1    & \textit{NumCls} \\
		\specialrule{.15em}{.2em}{.1em}
	\end{tabular}
	\caption{3D-ShuffleNetV2 architecture. Its' main building block is given in Fig. \ref{fig:blocks} (e) with stride 1 (left) and spatio temporal 2x downsampling (right). The number of channels ($c_1$, $c_2$, $c_3$, $c_4$) for different complexities are given in Table \ref{tab:shufflenetv2_ch}.}
	\label{tab:shufflenetv2_arch}
\end{table}

\begin{table}[t!]
    \centering
    \begin{tabular}{cccccc}
        \specialrule{.15em}{.0em}{.1em}
                                  & \multicolumn{5}{c}{\textbf{Output channels}} \\
                                  & \textbf{0.25x}  & \textbf{0.5x}  & \textbf{1.0x} & \textbf{1.5x} & \textbf{2.0x} \\
        \specialrule{.15em}{.1em}{.1em}
        \multicolumn{1}{c|}{$c_1$} & 32     & 48    & 116  & 176  & 244  \\
        \multicolumn{1}{c|}{$c_2$} & 64     & 96    & 232  & 352  & 488  \\
        \multicolumn{1}{c|}{$c_3$} & 128    & 192   & 464  & 704  & 976  \\
        \multicolumn{1}{c|}{$c_4$} & 1024   & 1024  & 1024 & 1024 & 2048 \\
        \specialrule{.15em}{.1em}{.0em}
    \end{tabular}
	\caption{The number of channels used in 3D-ShuffleNetv2 architecture for different levels of complexities.}
	\label{tab:shufflenetv2_ch}
\end{table}

\subsubsection{Comperative Analysis}

In this section, we compare the experimented architectures according to the number of layers, nonlinearities and skip connections. These design criteria plays an important role for the performance of the architectures. Comparison of the architectures are given in Table \ref{tab:model_comp}. For the number of layers, we counted the convolutional and linear layers. For the skip-connections, we have counted the addition or concatenation operations in the architectures. Finally, for the number of non-linearity, we have counted the ReLU operations in one inference time since it is the only non-linearity used for all the architectures. 

It is noticeable that comparatively earlier architectures (i.e. SqueezeNet and MobileNetV1) have smaller number of layers, non-linearity and skip-connections. On the other hand, recent resource efficient architectures (i.e. ShuffleNetV1, ShuffleNetV2 and MobileNetV2) are deeper, in the order of 50 layers and 30 non-linearity. Corollary, they require more skip connections in order to facilitate better gradient update mechanism.     

 \begin{table}[t!]
    \centering
    \begin{tabular}{lccc}
        \specialrule{.15em}{.2em}{.2em}
        \multirow{2}{*}{\textbf{Model}} & \multicolumn{3}{c}{\textbf{Number of}} \\ \cmidrule(lr){2-4}
                                                      & \textbf{layers} & \textbf{non-lin.} & \textbf{skip-con.}   \\
        \specialrule{.15em}{.2em}{.2em}
        3D-SqueezeNet                  & 18   & 18  & 4   \\
        3D-ShuffleNetV1                & 50   & 33  & 16  \\
        3D-ShuffleNetV2                & 51   & 34  & 16  \\
        3D-MobileNetV1                 & 28   & 27  & 0   \\
        3D-MobileNetV2                 & 53   & 35  & 10  \\
        \specialrule{.15em}{.2em}{.2em}
    \end{tabular}
    \caption{Comparison of resource efficient 3D architectures according to the number of layers, non-linearity and skip-connections.}
	\label{tab:model_comp}
\end{table}

\subsection{Training Details}

\textbf{Learning:} For the training of the architectures, Stochastic Gradient Descent (SGD) with
standard categorical cross-entropy loss is applied. For mini-batch size of SGD, largest fitting batch size is selected, which is usually in the order of 128 videos. The momentum, dampening and weight decay are set to 0.9, 0.9 and 1x10$^{-3}$, respectively. When the networks are trained from scratch, learning rate is initialized with 0.1 and reduced 3 times with a factor of 10$^{-1}$ when the validation loss converges. For the training of UCF-101 benchmark, we have used the pretrained models of Kinetics-600. We have frozen the network parameters and fine-tuned only the last layer. For fine-tuning, we start with a learning rate of 0.01 and reduced it two times after 30$^{th}$ and 45$^{th}$ epochs with a factor of 10$^{-1}$ and optimization is
completed after 15 more epochs.    

\textbf{Regularization:} Although Kinetics-600 and Jester are very large benchmarks and immune to over-fitting, UCF-101 still requires intensive regularization. Weight decay of 1x10$^{-3}$ is applied for all the parameters of the network. A dropout layer is applied before the final conv/linear layer of the networks. While dropout ratio is kept at 0.2 for Kinetics-600 and Jester, it is increased to 0.9 for UCF-101.


\textbf{Augmentation:} For temporal augmentation, input clips are selected from a random temporal position in the video clip. If the video contains smaller number of frames than the input size, loop padding is applied. For the input to the networks, always 16-frame clips are used. For Jester benchmark, it is critical to capture the full content of the gesture video in the selected input clip. Therefore, we have applied downsampling of 2 by selected 16 frames from 32 frames for Jester benchmark \cite{kopukluanalysis}. 

For spatial augmentation, we have selected a random spatial position from the input video. Moreover, we have selected a scale randomly from \{1, $\frac{1}{2^{1/4}}$, $\frac{1}{2^{3/4}}$, $\frac{1}{2}$\} in order to perform multi-scale cropping as in \cite{hara2018can}. For Kinetics-600 and UCF-101, input clips are flipped with 50\% probability. After the augmentations, input clip to the network has the size of 3 x 16 x 112 x 112 referring to number of input channels, frames, width and height pixels, respectively. 

\textbf{Recognition:} For Kinetics-600 and UCF-101, we select non-overlapping 16-frame clips from each video sample. Then center cropping with scale 1 is applied to each clip. Using the pretrained models, class scores for each clip is calculated. For each video, we average the scores of all clips. The class with the highest score indicates the class label of the video.

\textbf{Implementation:} Network architectures are implemented in PyTorch and trained with a single Titan Xp GPU.


%% file: Experiments.tex
\section{Experiments}

 \begin{table*}[t!]
    \centering
    \begin{tabular}{lccccccc}
        \specialrule{.15em}{.3em}{.3em}
        \multirow{2}{*}{\textbf{Model}} & \multirow{2}{*}{\textbf{MFLOPs}} & \multirow{2}{*}{\textbf{Params}} & \multicolumn{2}{c}{\textbf{Speed (cps)}} & \multicolumn{3}{c}{\textbf{Accuracy (\%)}}  \\ \cmidrule(lr){4-5} \cmidrule(lr){6-8}
                                        &              &              & \textbf{Titan XP} & \textbf{Jetson TX2} & \textbf{Kinetics-600} & \textbf{Jester} & \textbf{UCF-101} \\
        \specialrule{.15em}{.3em}{.3em}
        3D-ShuffleNetV1 0.5x       & 78           & 0.55M     & 398    & 69       & 35.51  & 89.23  & 64.39 \\
        3D-ShuffleNetV2 0.25x      & 116           & 0.83M     & 442    & 82       & 25.73  & 86.91  & 56.52 \\
        3D-MobileNetV1 0.5x        & 98           & 1.17M     & 290    & 57       & 31.74  & 87.61  & 62.17 \\
        3D-MobileNetV2 0.2x        & 63           & 0.96M     & 357    & 42       & 24.14  & 86.43  & 55.56 \\
        \specialrule{.15em}{.3em}{.3em}
        3D-ShuffleNetV1 1.0x       & 199          & 1.52M     & 269    & 49       & 45.31  & 92.27  & 76.00 \\
        3D-ShuffleNetV2 1.0x       & 195          & 1.91M     & 243    & 44       & 46.10  & 91.96  & 77.90 \\
        3D-MobileNetV1 1.0x        & 241          & 3.91M     & 164    & 31       & 40.07  & 90.81  & 70.95 \\
        3D-MobileNetV2 0.45x       & 177          & 1.40M     & 203    & 19       & 36.47  & 90.21  & 68.31 \\
        \specialrule{.15em}{.3em}{.3em}
        3D-ShuffleNetV1 1.5x       & 347          & 2.92M     & 204    & 31       & 52.75  & 93.12  & 81.73 \\
        3D-ShuffleNetV2 1.5x       & 291          & 3.16M     & 186    & 34       & 52.05  & 93.16  & 82.32 \\
        3D-MobileNetV1 1.5x        & 429          & 8.22M     & 116    & 19       & 48.24  & 91.28  & 76.00 \\
        3D-MobileNetV2 0.7x        & 325          & 2.05M     & 130    & 13       & 45.59  & 93.34  & 77.32 \\
        \specialrule{.15em}{.3em}{.3em}
        3D-ShuffleNetV1 2.0x       & 538          & 4.76M     & 161    & 24       & 56.84  & 93.54  & 84.96 \\
        3D-ShuffleNetV2 2.0x       & 438          & 6.64M     & 146    & 26       & 55.17  & 93.71  & 83.32 \\
        3D-MobileNetV1 2.0x        & 662          & 14.10M    & 88     & 15       & 48.53  & 92.56  & 76.18 \\
        3D-MobileNetV2 1.0x        & 561          & 3.12M     & 93     & 9        & 50.65  & 94.59  & 81.60 \\
        3D-SqueezeNet              & 926          & 2.15M     & 682    & 46       & 40.52  & 90.77  & 74.94 \\
        \specialrule{.15em}{.3em}{.3em}
        ResNet-18                  & 8323          & 33.36M     & 334     & 17       & 57.65  & 93.34  & 80.09 \\
        ResNet-50                  & 9835          & 44.54M     & 183     & 11       & 63.00  & 93.70  & 88.92 \\
        ResNet-101                 & 13664         & 83.58M     & 142     & 8       & 64.18  & 94.10  & 87.02 \\
        ResNeXt-101                & 9652          & 48.75M     & 122     & 7       & 68.30  & 94.89  & 89.08 \\
        I3D \cite{carreira2018short} & 111331       & 12.70M     & ---     & ---       & 71.90  & ---  & --- \\
        \specialrule{.15em}{.3em}{.3em}
    \end{tabular}
    \caption{Comparison of resource efficient 3D architectures over video classification accuracy, number of parameters and speed on two different platforms and four levels of computation complexity. The calculations of MFLOPs, parameters and speeds are done for Kinetics-600 benchmark. For speed calculations (clips per second (cps)), the used platforms are Titan Xp and Jetson TX2; and the batch size is set to 8. All models takes 16 frames input with 112 x 112 spatial resolution except for I3D, which takes 64 frames input with 224 x 224 spatial resolution.}
	\label{tab:comparison}
\end{table*}

In this section, we first explain the experimented datasets. Then, we discuss about the achieved results for the experimented network architectures together with their run-time performance on both NVIDIA Titan Xp and Jetson TX2 embedded system. 

\subsection{Datasets}

\noindent $\bullet$ \textbf{Kinetics-600 dataset} is an extension of Kinetics-400 dataset, which contains 600 human action classes, with at least 600 video clips for each action. Each clip is approximately 10 seconds long and is taken from a different YouTube video. There are in total 392,622 training videos. For each class, there are also 50 and 100 validation and test videos, respectively. Since the labels for the test set is not publicly available, we have conducted our experiments on the validation set.

We selected Kinetics-600 benchmark in order to evaluate the capacity of the experimented networks. It is very rare that a real-life application tries to classify 600 different classes. However, these kind of very large-scale datasets are very useful to evaluate the capacity of the networks to learn. Although it is still necessary to capture the motion patterns in the video, the network should especially capture the spatial content in order to identify the correct class label of the video. For example, there are 9 different "eating something" classes where "something" is one of "burger, cake, carrot, chips, doughnut, hotdog, ice cream, spaghetti, watermelon". Although "eating" action is same for all these, the true label can only be identified when the network captures discriminative features of what is being eaten. 

\noindent $\bullet$ \textbf{Jester dataset} is currently the largest available hand gesture dataset. In each video sample of the dataset, a person performs pre-defined hand gestures in front of a laptop camera or webcam. There are in total 148,092 gesture videos under 27 classes. The dataset is divided into three subsets: training set (118,562 videos), validation set (14,787 videos), and test set (14,743 videos). Since the labels for test set is not publicly available, we have conducted our experiments on the validation set.

Unlike Kinetics-600 benchmark, in Jester dataset, spatial content of the all video samples are same: A person sitting in front of a camera performs a hand gesture from almost the same distance. Moreover, the selection of classes are more focused on the movement of the hand. That is why, Jester benchmark is suitable to inspect the ability of the networks in capturing motion patterns.

\noindent $\bullet$ \textbf{UCF101 dataset} is an action recognition dataset of realistic action videos, collected from YouTube. It consists of 101 action classes, over 13k clips and 27 hours of video data. Compared to Kinetics-600 and Jester datasets, UCF-101 contains very little amount of training videos, hence prone to over-fitting. For the evaluation of UCF-101 dataset, we have used only split-1. We selected UCF-101 benchmark in order to inspect the applicability of transfer learning for the experimented network architectures. 


\subsection{Results}

In this section, we elaborate on our findings in the experiments that we have conducted for 5 different network architectures, 4 levels of complexity (except for SqueezeNet) on 3 different benchmarks. Moreover, runtime performance of the models are evaluated on 2 different platforms, namely Titan XP and Jetson TX2 embedded system. According to the results in Table \ref{tab:comparison}, the following conclusions can be inferred:

\vspace{0.1cm}
\textbf{Accuracy:}

\vspace{0.05cm}
\textbf{(i)} The deeper architectures (3D-ShuffleNet, 3D-ShuffleNetV2, 3D-MobileNetV2) achieve better results compared to shallower architectures (3D-SqueezeNet, 3D-MobileNetV1). Accordingly, resource efficient 3D CNNs should not be designed too shallow in order to save complexity. 

\vspace{0.05cm}
\textbf{(ii)} Motion patterns are better captured with depthwise convolutions. Since depthwise convolutions have kernels of 3x3x3, they can capture relations in depth dimension together with spatial dimension. The main building block of 3D-MobileNetV2 is the inverted residual block, which expands the number of channels to the input of depthwise convolution layers with an expansion ratio. Therefore, it contains more depthwise convolution filters compared to other architectures. Consequently, it achieves by far best performance in Jester benchmark, although it has inferior results in Kinetics-600 and UCF-101 benchmarks.

\vspace{0.05cm}
\textbf{(iii)} All models showed comparatively similar performance on both Kinetics-600 and UCF-101 datasets. This shows transfer learning is a valid approach for resource efficient 3D CNNs since there is a direct correlation between model performances on these two datasets.

\vspace{0.1cm}
\textbf{Complexity level:}

\vspace{0.05cm}
\textbf{(iv)} There is a severe performance degradation if the networks are scaled with very small $width\_multiplier$ in order to satisfy the required computational complexity. For example, in the first block of the Table \ref{tab:comparison}, we can see that 3D-MobileNetV2 0.2x and 3D-ShuffleNetV2 0.25x achieve 5-9\% worse than 3D-ShuffleNetV1 0.5x and 3D-MobileNetV1 0.5x in Kinetics-600 benchmark. Capacity of the models degrades severely as the $width\_multiplier$ gets smaller, especially when it is less than 0.5.  We can see the same pattern on all three benchmarks that we have experimented. 

\vspace{0.05cm}
\textbf{(v)} The main design criteria of the 3D-SqueezeNet is to save number of parameters, not computations. Therefore it has the smallest number of parameters at the highest complexity level. However, it also has around 300 million more FLOPs compared to other architectures since it does not make use of depthwise convolutions.

\vspace{0.1cm}
\textbf{Runtime performance:}

\vspace{0.05cm}
\textbf{(vi)} Although the network architectures contain similar FLOPs, some architectures are much faster than others. As highlighted by \cite{ma2018shufflenet}, this is due to several other factors affecting speed such as memory access cost (MAC) and degree of parallelism, which are not taken into account by FLOPs.

\vspace{0.05cm}
\textbf{(vii)} 3D-SqueezeNet is the only architecture that does not make use of depthwise convolutions, hence contains highest FLOPs. However, surprisingly it has the highest runtime performance. This is due to the latest CUDNN \cite{chetlur2014cudnn} library which is specifically optimized for standard convolutions. Similar results can also be observed with ResNet and ResNeXt architectures.

\vspace{0.05cm}
\textbf{(viii)} Runtime performance heavily depends on the hardware that the network architecture is running. For example, for the highest two complexity levels, 3D-ShuffleNetV1 is the faster than 3D-ShuffleNetV2 on GPU, whereas 3D-ShuffleNetV2 achieves higher runtime than 3D-ShuffleNetV1 on Jetson TX2.

\vspace{0.1cm}
\textbf{State-of-the-art comparison:}

\vspace{0.05cm}
\textbf{(ix)} Architectures with more parameters and FLOPs like ResNets, ResNeXt-101 and I3D achieve generally better results for datasets measuring the capacity of the tested architectures like Kinetics dataset as evaluated and shown in Table \ref{tab:comparison}. However, network design makes a huge difference. For example, 3D-ShuffleNetV1 2.0x achieves similar performance with ResNet-18, although ResNet-18 requires 7 times more parameters and 15 times FLOPs .  

\vspace{0.05cm}
\textbf{(x)} The architecture design should be done according to the given task. As inverted residual block excels at capturing dynamic motions, 3D-MobileNetV2 1.0x achieves better results than much wider and deeper ResNet-101 (around 20 times more parameters and FLOPs) at Jester benchmark.

%% file: Conclusion.tex
\section{Conclusion}

In recent years, the research in action recognition has mostly focused on obtaining the best accuracy by generating deep and wide CNN architectures. However, real-world applications require resource efficient architectures that take runtime, memory and power budget into account. Recently, several resource efficient 2D CNN architectures have been proposed. However, there is a lack of architectures for 3D counterparts. This work aims to fill this research gap. 

The proposed architectures are generated by implementing the 3D versions of Squeezenet, MobileNet, MobileNetV2, ShuffleNet, ShuffleNetV2 architectures for 4 different complexity levels. The performance of these architectures have been evaluated using 3 different benchmarks, which are selected according to analyze models' capacities, how well the models capture the motion and the applicability of transfer learning for each model.

According to the analysis for 4 different complexity levels, the results show that these resource efficient 3D CNN architectures provide considerable classification performances. Using the $width\_multiplier$, the capacity of the architectures can be modified flexibly. The results on Jester benchmark show that depthwise convolutions are very good at capturing motion patterns. Moreover, nearly all models run in real-time both at Titan XP and Jetson TX2. As the results proved the applicability of transfer learning, these architectures can be used for other real-world applications by using pretrained models.


%% file: Acknowledgements.tex
\section*{Acknowledgements}
We give our special thanks to Stefan H\"ormann for his assistance to this work. We also gratefully acknowledge the support of NVIDIA Corporation with the donation of the Titan Xp GPU and Jetson TX2 Development Kit used for this research.